# Does Explainability Survive Data Drift?

**Samuel Ozechi***

## Abstract

Model performance monitoring is a standard practice in machine learning deployments. Detection performance is tracked continuously, and model decay is expected as the relationship between the feature and target variables degrades, a phenomenon known as concept drift. Explanation fidelity, however, is rarely monitored with the same discipline, even in domains such as financial systems, healthcare, and other regulated environments where explanations are required for governance purposes. This paper investigates whether explanations can decay under data drift, even when the feature–target relationship remains stable, and whether explanations produced before drift occurs remain faithful to the decisions of the model that replaces them. Using the IEEE-CIS Transaction Fraud Detection dataset, we confirm that the dataset exhibits statistically significant covariate shift without concept drift, thereby isolating input distributional change from changes in the feature-target relationship. Local explanations are generated with ExIFFI and evaluated at three levels: path validity, structural behaviour, and fidelity under controlled intervention. Results show that while prior explanations retain substantial decision-relevance to a retrained model, they are consistently less faithful than newly generated explanations, with the gap stable across windows rather than compounding with drift. The study shows that explanation fidelity requires its own monitoring, that structural stability of explanations does not guarantee functional fidelity, and that explanations should be treated as artifacts tied to the model that produced them.

## 1. Introduction

Machine learning-based anomaly detection has become an important component of systems where abnormal behaviour must be identified from large volumes of data [6], [20].

In financial transactions, for example, anomaly detection models can identify potentially fraudulent activities within highly heterogeneous and continuously evolving transaction streams [7].

However, the domains in which detection models operate are rarely stationary. Transaction and user behaviours, merchant properties, attack strategies, and other attributes of the data can change over time.

Consequently, a model that performs well in development, may encounter a substantially different data distribution after deployment.

This behaviour is commonly described as data drift, which refers to a change in the distribution of input features $P(X)$, while the relationship between features and the target $P(Y|X)$ may remain stable [9], [18].

Conventional monitoring typically focuses on whether the model's predictive performance remains acceptable as new observations arrive. Metrics such as F1 score and area under the curve (AUC) can reveal whether the model's ability to identify anomalous observations changes over time.

Such metrics are essential, particularly in highly imbalanced applications such as fraud detection, where precision-recall analysis can be more informative than ROC-based evaluation [21], but they describe only one aspect of model behaviour.

A model may continue to detect anomalies while the ranked features flagged as responsible no longer accurately reflect the detector's true decision mechanism.

Conversely, changes in the model's explanation structure do not necessarily imply a decline in detection performance. Detection and explanation quality are distinct properties that require separate measurement, and this separation becomes

*Corresponding author: Samuel Ozechi

**Email address:**

ozechisamuel@gmail.com (Samuel Ozechi

particularly when explanations are needed to justify why an observation was flagged as anomalous.
Explainability methods are increasingly used to associate individual model outputs with input features, providing users with a representation of the factors that contributed to a particular prediction or anomaly score [16], [19].
The broader literature on explainable anomaly detection has also highlighted the need for explanations that help users understand the factors underlying anomalous behaviour [13], [17]. Model-specific explanation approaches have therefore been developed for anomaly detectors, including feature importance methods designed to expose the internal decision structure of Isolation Forest models [2], [5].
Yet an explanation may identify a set of features that remains consistent across observations, even if those features have limited influence on the model's actual output.
This distinction motivates an important separation between explanation stability and explanation fidelity.
Stability concerns whether an explanation or feature ranking remains similar across temporal windows, whereas fidelity concerns whether it accurately reflects the behaviour of the model being explained [3], [23].
Our research therefore treats stability as a useful diagnostic, but not as sufficient evidence that an explanation accurately represents the model's decision.
Ideally, if the feature–target relationship $P(Y|X)$ changes, the features that explain its outputs, or their contributions, are expected to change accordingly; this is known as concept drift.
This paper focuses on the complementary case when $P(Y|X)$ is stable, but the input distribution $P(X)$ shifts. It is not obvious whether explanations that were faithful under the training distribution remain faithful under the drifted input distribution, or whether the relationship between explanation and decision silently degrades. This motivates the central question of the study: Does explainability survive data drift?

## 2. Related Work and Contributions

Anomaly detection has been extensively studied as a means of identifying observations that deviate from the dominant structure of a dataset, with applications spanning fraud detection, cybersecurity, industrial monitoring, and other high-risk domains [1], [6], [11].
Isolation Forest (IF) is particularly relevant to this domain because it provides an efficient unsupervised mechanism for identifying observations that can be isolated using random partitioning of the feature space [14].
Subsequent works have extended the original axis-aligned formulation to address limitations associated with axis-parallel partitions, including Extended Isolation Forest (EIF), which uses randomly oriented hyperplanes [10].
These developments also motivated research into making isolation-based detectors interpretable. DIFFI introduced a model-specific feature-importance mechanism that exploits the structure of Isolation Forest paths to identify features contributing to the isolation of observations [5].
ExIFFI subsequently extended this towards explanations for Extended Isolation Forest and related isolation based models, providing local and global feature-importance information derived from the model's internal decision structure [2].
Such approaches are particularly relevant to the present study because they allow explanations to be evaluated against the actual structure of the anomaly detector rather than treating the detector as an entirely opaque predictive function.
More broadly, XAI has produced a range of model-agnostic and model-specific approaches for attributing model outputs to input features, including LIME and SHAP [19].
The distinction between explanation generation and fidelity has motivated various approaches to explanation evaluation, including approaches in which the most important features are perturbed or removed, and the resulting change in model output is measured [8], [17], [23].
However, the question of whether such explanations remain faithful when the input distribution changes over time is less directly established.
This creates an important intersection between two bodies of research. The concept-drift literature has primarily focused on detecting changes in data-generating relationships and maintaining performance through model adaptation [9], while explanations have largely been evaluated with respect to static settings [19].
Similarly, work on interpretable Isolation Forests has focused primarily on deriving feature importance from the detector's internal structure. These perspectives do not by themselves establish whether an explanation that is faithful for a model under one input distribution remains faithful when the model encounters a different input distribution.
Consequently, detection performance, explanation structure, and decision-level explanation fidelity represent distinct properties that should be evaluated separately.
This study therefore provides a unified empirical framework for investigating whether explanation fidelity survives input distributional change and model adaptation.
Its contributions:

1. Establishes an empirical evaluation setting defined by substantial covariate shift without statistically detectable conditional concept drift..
2. Provides a decision-level framework for validating explanation fidelity.
3. Evaluates ExIFFI explanations for a static Isolation Forest under data drift, tracking how explanation structure and fidelity evolve across windows.
4. Analyzes cross window explanation transferability, thereby establishing whether prior explanations still describe current decisions.
5. Demonstrates that model performance, explanation structure, and fidelity are distinct properties of an anomaly detection system that require independent monitoring.

## 3. Research Questions

The research questions are organised around four levels of analysis: performance, explanation structure, fidelity, and their relationships.

**RQ1 (Detection Performance):** How does the detection performance of a static Isolation Forest model change across temporally ordered future windows as the input distribution drifts?

This sets the reference against which explanation behaviour is compared. If detection performance degrades, explanation degradation should be interpreted against that backdrop. If detection performance remains stable while explanation degrades, the degradation becomes a finding of independent interest.

**RQ2 (Explanation Structure):** How do the structural properties of explanations change across future windows as the input distribution drifts?

A change in the structural properties of local explanations indicates that the distribution of attributed importance across features is shifting, but it does not establish whether the explanations remain faithful to the model's decisions.

This question separates descriptive change from decision level fidelity.

**RQ3 (Decision-Level Fidelity):** Do the features identified by ExIFFI explanations actually account for the model's decisions, and does fidelity change across future windows?

If explanations remain faithful to the model's decisions across windows, then monitoring detection performance is sufficient for the operational reliability of explanations.

If fidelity degrades, then fidelity itself must be tracked as a separate property, and this changes how deployment and compliance are handled in systems where explanations are relied upon.

**RQ4 (Explanation Transferability):** Do explanations of a previous model remain faithful to the decision logic of an updated model?

If prior explanations remain faithful to an updated model, then the explanations are transferable and do not need to be regenerated after each update.

If not, then explanation fidelity is tied to the model that generated the explanations, and any requirement for accurate explanations implies that they must be regenerated after model updates.

## 4. Methods

The transactions are ordered chronologically, with the earliest 60% of observations used as the initial training period and the remaining 40% divided into four equal, chronologically ordered evaluation windows (W1–W4).

All preprocessing operations, including feature selection, missingness filtering, and imputation, are fitted using the initial training period only and then applied unchanged to all subsequent windows. This prevents information from future observations from entering the modelling or explanation pipeline.

Each evaluation window contains 59,054 transactions and fraud prevalence varies across the four windows, providing temporal variation in class balance alongside changes in the observed feature distributions.

### 4.1. Drift Analyses

Before evaluating explanations, the nature of the input distributional changes in the dataset is verified. Data drift is assessed through changes in the input distribution $P(X)$.

For each feature, the training-period distribution is compared with the distribution observed in each future window using the two-sample Kolmogorov–Smirnov (KS) test and Jensen Shannon (JS) divergence.

Also, concept drift is tested directly as it concerns changes in conditional relationship $P(Y|X)$, rather than just changes in inputs $P(X)$ alone. A classifier-based conditional drift test is performed independently for each future window using the training period and the corresponding evaluation window. The test compares the ability to distinguish between the two populations using the feature representation $X$ alone with the ability obtained when the joint information in $X$ and $Y$ is additionally available [15].

An observed improvement in discrimination from including $Y$ beyond $X$ provides evidence that the relationship between the feature distribution and the target differs between the two periods.

Statistical significance is assessed using 200 permutations per window, and the resulting p-values are adjusted across windows using the Benjamini–Hochberg false discovery rate (BH-FDR) procedure.

As a complementary test, a Random Forest fraud classifier is trained using the training period and evaluated on each future window. Predictive performance on the training period is compared with performance on each future window, with the observed change in the ROC-AUC when assessed against a permutation based null distribution, with BH-FDR correction applied across the windows.

Together, these analyses utilize statistical frameworks to distinguish distributional change in $P(X)$ from conditional concept drift in $P(Y|X)$.

### 4.2. Model Conditions

One Isolation Forest is trained on the training period and applied unchanged to every future window. This model is never trained on the drifted distribution, and its decision process is fixed throughout the study.

Four other Isolation Forests are then trained on each future window (one per future window), and on that window's data only. Each retrained model is then applied to all future windows for performance and explanation analyses.

Comparing the two model conditions (static and retrained)

isolates the effect of adaptation from the effect of drift.
The static model tests whether the explanations of a fixed decision process remain valid as the input distribution shifts, while the retrained models test whether adapting the decision process to the shifted distribution preserves explanation and its fidelity.
Both model conditions use the same hyperparameters and preprocessing. The retrained models differ from the static model only in the data they are trained on.

## 4.3. Explanation Structure and Fidelity

Local explanations are generated using ExIFFI, a model specific feature-importance method developed for Isolation Forest and its extended variants [2]. ExIFFI derives feature contributions from the internal decision paths followed by an observation through the isolation trees.
For a standard axis-aligned Isolation Forest, each internal node splits on exactly one feature, so the split-direction vector reduces to a single basis vector. ExIFFI assigns the node's contribution entirely to that feature. The path weight for a node $k$ is:

$$\lambda_k(x) = \frac{|X_k|}{|X_{k+1}|}$$

Where $X_k$ is the number of training samples reaching the parent node and $X_{k+1}$ is the number reaching the child. Contributions are accumulated across all trees and all path nodes for the observation, yielding a raw per-feature vector. The vector is normalised to sum to one for inspection, so that importance values are directly comparable across inputs.
For each explanation vector, the mass, entropy, and features with non-zero importance are computed.
These metrics describe the structure of the explanation, but do not establish whether the explanation is faithful to the model’s decision [8].
The fidelity analysis is measured by an intervention. For an observation $x$ and its ExIFFI explanation, the top-k features are identified by importance. The intervention replaces those top-k features with training-period medians and leaves all remaining features unchanged, producing a perturbed version of the observation ($x'$).
The model is then re-scored on $x'$, and the change relative to the original score is measured. If the top-k features identified are actually the ones that drive the decision, then replacing them should change the model's output substantially. If the score is essentially unchanged, the identified features are not materially responsible for the model’s decision.
For an observation $x$ and its intervention $x'$, the relative score change at $k$ is defined as:

$$\text{Relative Score Change} = \frac{|score(x') - score(x)|}{|score(x) + \epsilon}$$

The primary reported value is $k = 5$ chosen as a mid-range value that is small enough to test precision and large enough to remain stable against sampling noise.

# 5. Results

By isolating input distribution shifts from changes in the underlying feature-label relationship, the impact of data drift on local explanation behaviour and fidelity is examined for Isolation Forest under static and retrained conditions..

## 5.1. Distributional Changes in the Dataset

A Random Forest classifier trained on the training period and evaluated on each future window shows no statistically significant performance drop after BH-FDR correction.
The absence of a significant performance drop suggests that the conditional relationship between the features and label is stable across the evaluated periods, and concept drift is not statistically detectable in the dataset.

**Table 5.1.1. Performance-drop test for concept drift**

| Window | Samples | Fraud rate | Performance drop | BH-FDR adjusted p |
|---|---|---|---|---|
| **W1** | 59,054 | 4.32% | −0.0272 | 1.000 |
| **W2** | 59,054 | 3.49% | −0.0128 | 1.000 |
| **W3** | 59,054 | 3.13% | +0.0041 | 0.911 |
| **W4** | 59,054 | 3.75% | −0.0094 | 1.000 |

However, stable performance alone does not rule out concept drift. The conditional distribution test shows that including the fraud labels in a classifier trained to differentiate training period observations from the future window observations does not improve it.
For each window, the improvement in accuracy from adding $Y$ to the feature set ranges is indistinguishable from the permutation null at every window.
If the conditional distribution $P(Y|X)$ had changed, knowing $Y$ would help the classifier better identify which window an observation came from. It does not. These two lines of evidence point to the absence of concept drift in the dataset.

**Table 5.1.2. Conditional distribution test**

| Window | Accuracy | Accuracy X, Y | Null Mean | BH-FDR adjusted p |
|---|---|---|---|---|
| **W1** | 0.68 | 0.68 | +0.0041 | 0.806 |
| **W2** | 0.69 | 0.70 | +0.0051 | 0.806 |
| **W3** | 0.70 | 0.70 | +0.0018 | 0.806 |
| **W4** | 0.70 | 0.70 | -0.0016 | 0.806 |

In contrast, a clear majority of features exhibit significant distributional change across all windows, comprising 202 of 376 features (53.7%) in W1, 264 (70.2%) in W2, 231 (61.4%) in W3, and 244 (64.9%) in W4. The maximum KS statistic ranges from 0.157 (W1) to 0.225 (W3), indicating that at least one feature in each window has substantial distributional separation from the training period. The mean JS divergence rises monotonically across the first three windows and remains elevated at W4.

**Table 5.1.3. Covariate shift per window**

| Window | Significant KS (%) | Max KS Statistic | Mean JS divergence | Median JS Divergence |
|---|---|---|---|---|
| **W1** | 53.7 | 0.157 | 0.00137 | 0.00044 |
| **W2** | 70.2 | 0.193 | 0.00172 | 0.00063 |
| **W3** | 61.4 | 0.225 | 0.00177 | 0.00049 |
| **W4** | 64.9 | 0.201 | 0.00179 | 0.00032 |

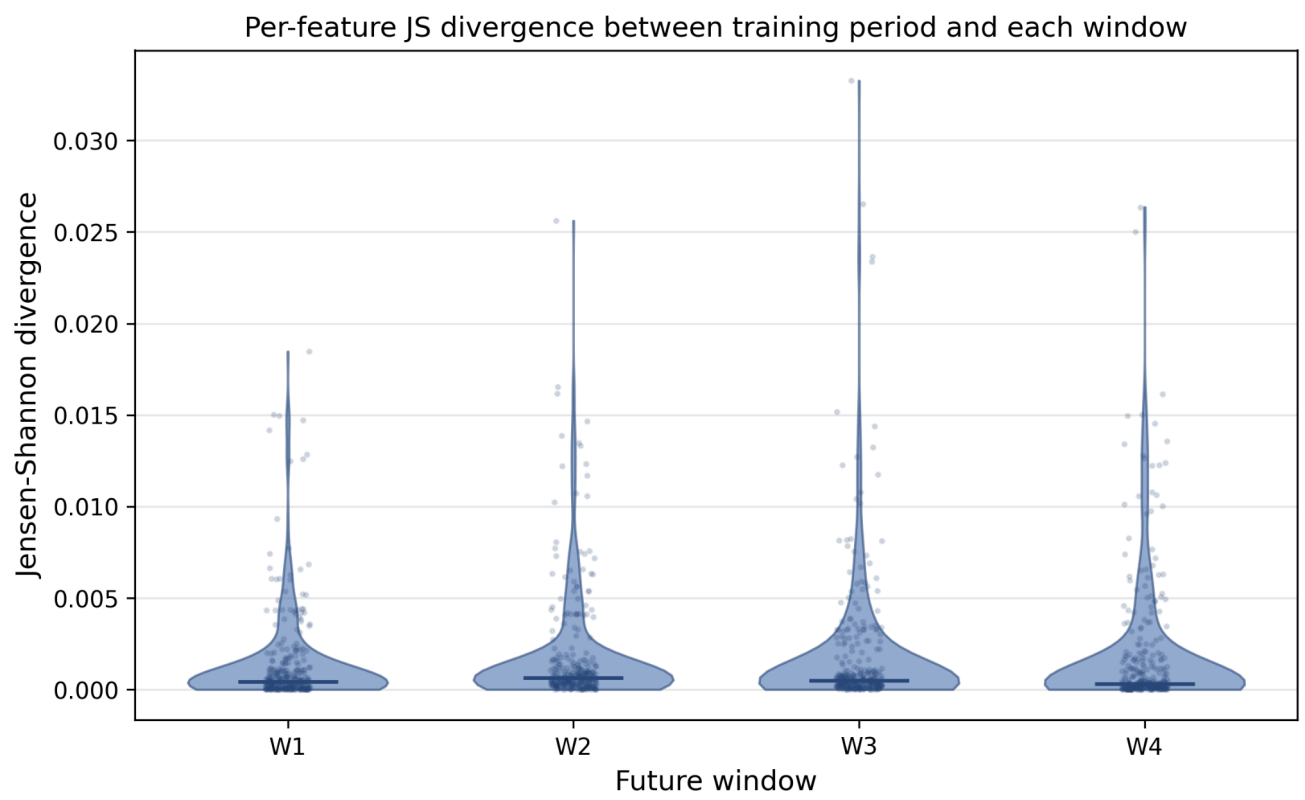

*Figure 5.1.1. Feature distribution shift across windows.*

Fraud rates fluctuate across windows, dipping from 4.32% in W1 to 3.13% in W3 before rebounding to 3.75% in W4.
This corresponds to label shift of moderate magnitude in $P(Y)$, rather than necessarily a change in $P(Y|X)$ [18].

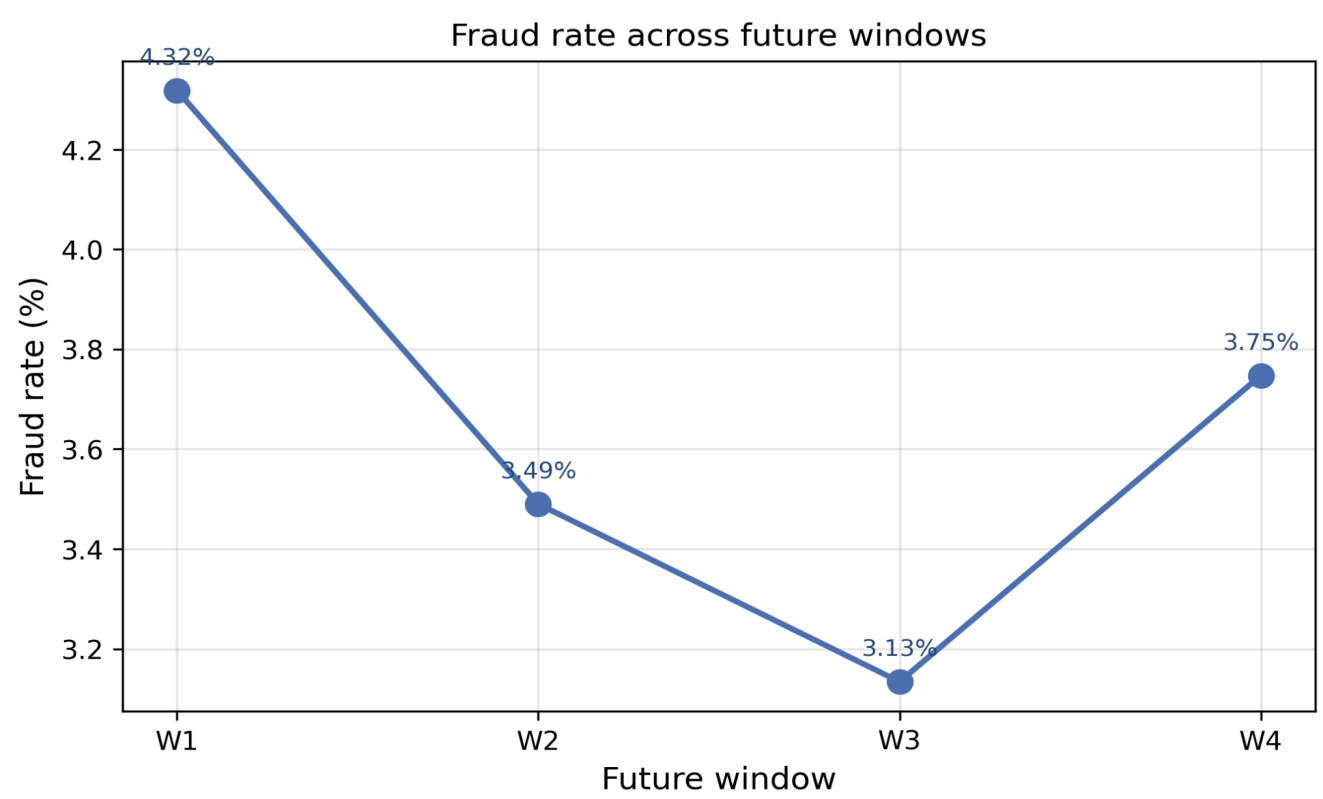

*Figure 5.1.2. Fraud rate across future windows.*

This characterisation establishes the setting for the rest of the study, as any observed changes in explanation behaviour are therefore interpreted as occurring under covariate shift.

## 5.2. Detection Benchmark

The Isolation Forest anomaly detection evaluation reveals three consistent trends regarding detection capability across temporal evaluation windows W1 to W4 .
First, all evaluated models exhibit monotonic performance decline over time. Between W1 and W4, F1 scores drop by 24% to 30% across models, while Precision Recall-AUC (PR-AUC) suffers a substantial reduction of 45% to 49%. For the static baseline model specifically, F1 drops from 0.27 to 0.20 (a 24% decline), alongside a 46% drop in PR-AUC. This performance drop is most pronounced between W2 and W3, coinciding with a reduction in the class distribution, where the observed fraud rate reduces from 3.49% to 3.13%.
Secondly, retraining consistently mitigates performance loss relative to the static baseline across all evaluation windows, although it fails to halt temporal degradation entirely.
Retrained models sustain higher performance, maintaining an F1 advantage of 6.4% to 9.5% and a PR-AUC advantage of 1.2% to 8.8% over the static model across windows.
At W1, the optimal F1 score reaches 0.30 under retrained model W3 compared to 0.27 for the static baseline; at W4, the peak F1 score is 0.22 under retrained model W3 against 0.20 for the static model. Crucially, each retrained variant still follows a downward trajectory as it is evaluated across subsequent windows.

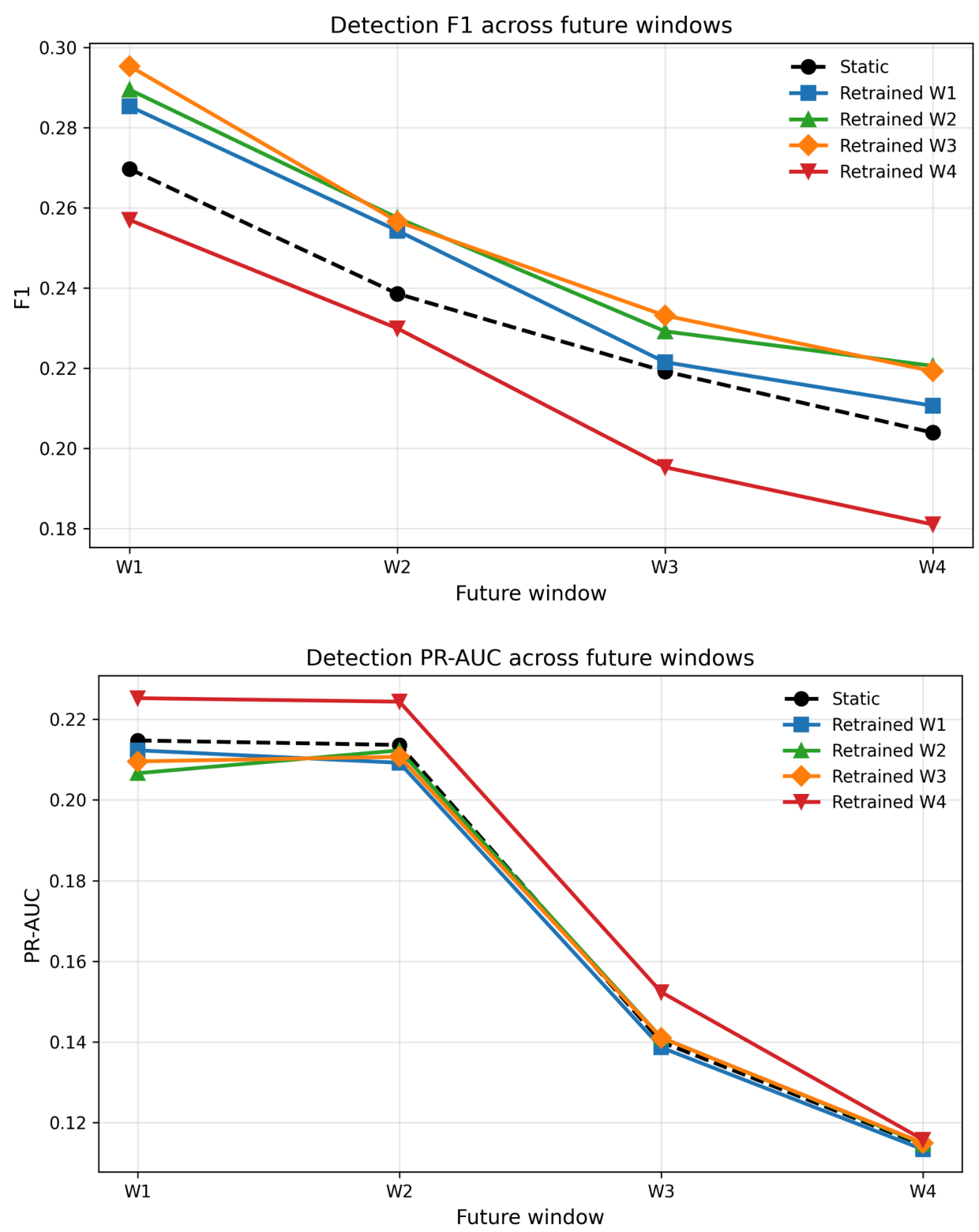

*Figure 5.2. Detection trajectories across future windows*

Conversely, training on parallel windows fails to optimize F1 -scores, with off-diagonal models outperforming in two windows, whereas the W4-trained model achieves superior PR-AUC across all windows, suggesting that W4 captures a broader, more representative fraud distribution.

## 5.3. Explanation Structure

The structural properties of the explanation vectors are evaluated and compared to examine how feature importance distributions evolve under data drift.

Across all windows, explanation concentration exhibits slight but consistent declines over time, accompanied by a corresponding rise in entropy. For the static baseline, the primary concentration metric (top-5 mass) drops from 0.18 at W1 to 0.17 at W4, while entropy rises from 5.21 to 5.3 nats, expanding the effective feature count from 206 to 222.

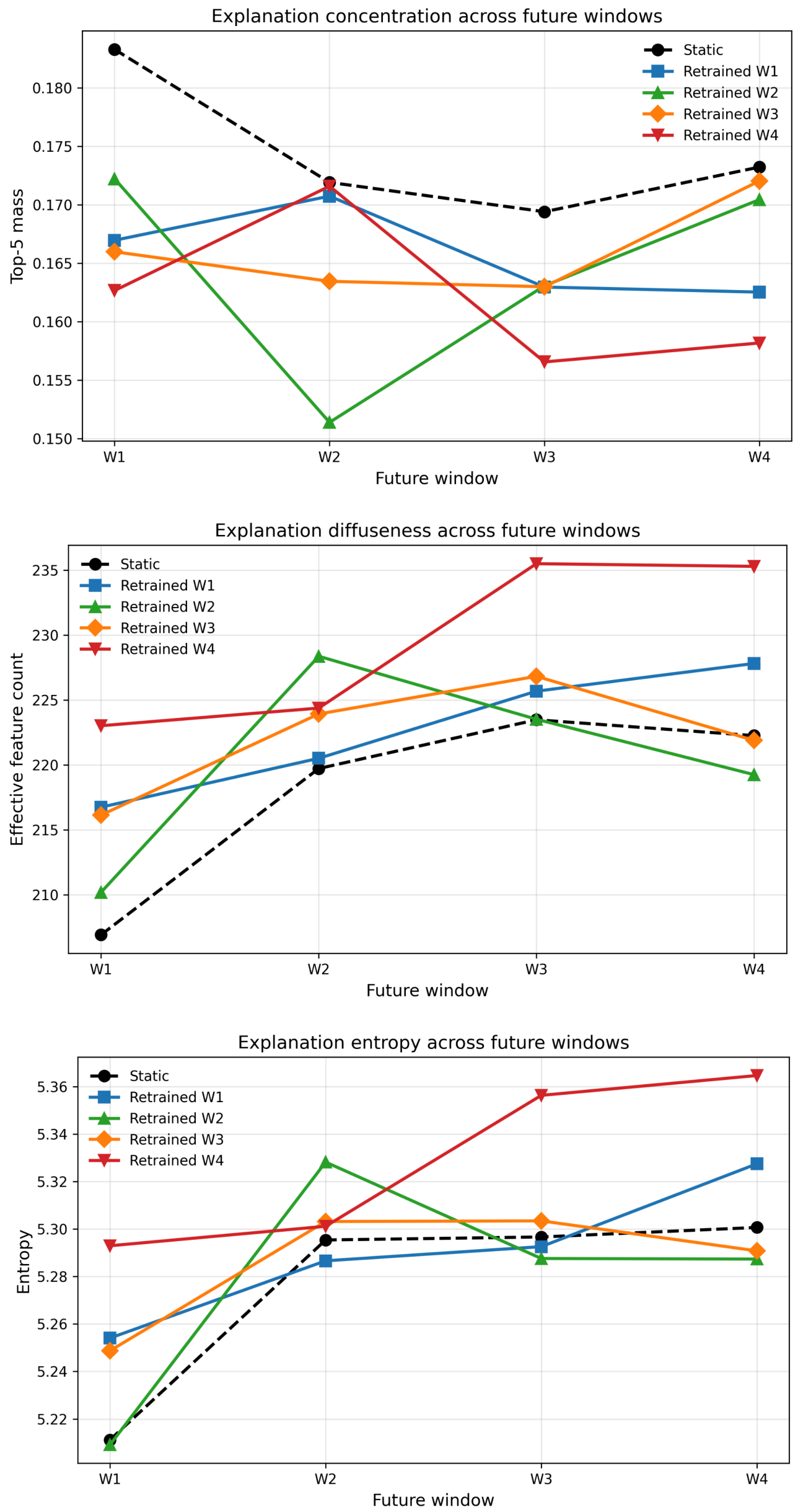

*Figure 5.3* *Explanation structure across windows*

Also, the static model consistently maintains the highest degree of explanation concentration and the lowest entropy across all windows. At W1, the static model's top-5 mass of 0.18 surpasses all retrained counterparts (ranging from 0.15 to 0.17). This dominance persists at W4, where its top-5 mass (0.17) remains higher than any retrained variant (0.15 to 0.17). Training across the entire historical baseline yields explanations that concentrate attribution weight onto fewer features compared to models trained on localized temporal windows.

The effective feature count, which factors feature importance weights rather than binary presence, serves as the true metric for assessing attribution diffuseness.

Raw non-zero feature counts exhibit near-saturation across every combination, assigning positive attribution to between 367 and 362 of the 376 available features. Consequently, unweighted counts do not discriminate between experimental conditions.

## 5.4. Explanation Fidelity

The central fidelity validation compares the ExIFFI-selected features against a random feature intervention that applies the identical operation but selects $k$ features uniformly at random.

This follows the broader principle of evaluating explanations through their effect on the model's output under controlled interventions, rather than relying solely on the stability of the resulting feature rankings [8], [19].

The results show that the ExIFFI-selected features condition produces a substantially larger mean relative score change than the random control at every $k$.

At $k = 1$, the difference exceeds two thousand percent; the ratio of ExIFFI to random is 23.8×. The gap narrows as k increases, falling to 12.2× at k=20, which is expected: larger random subsets have a higher probability of including a decision-relevant feature by chance, so the random control improves with $k$ while the ExIFFI advantage is partially absorbed.

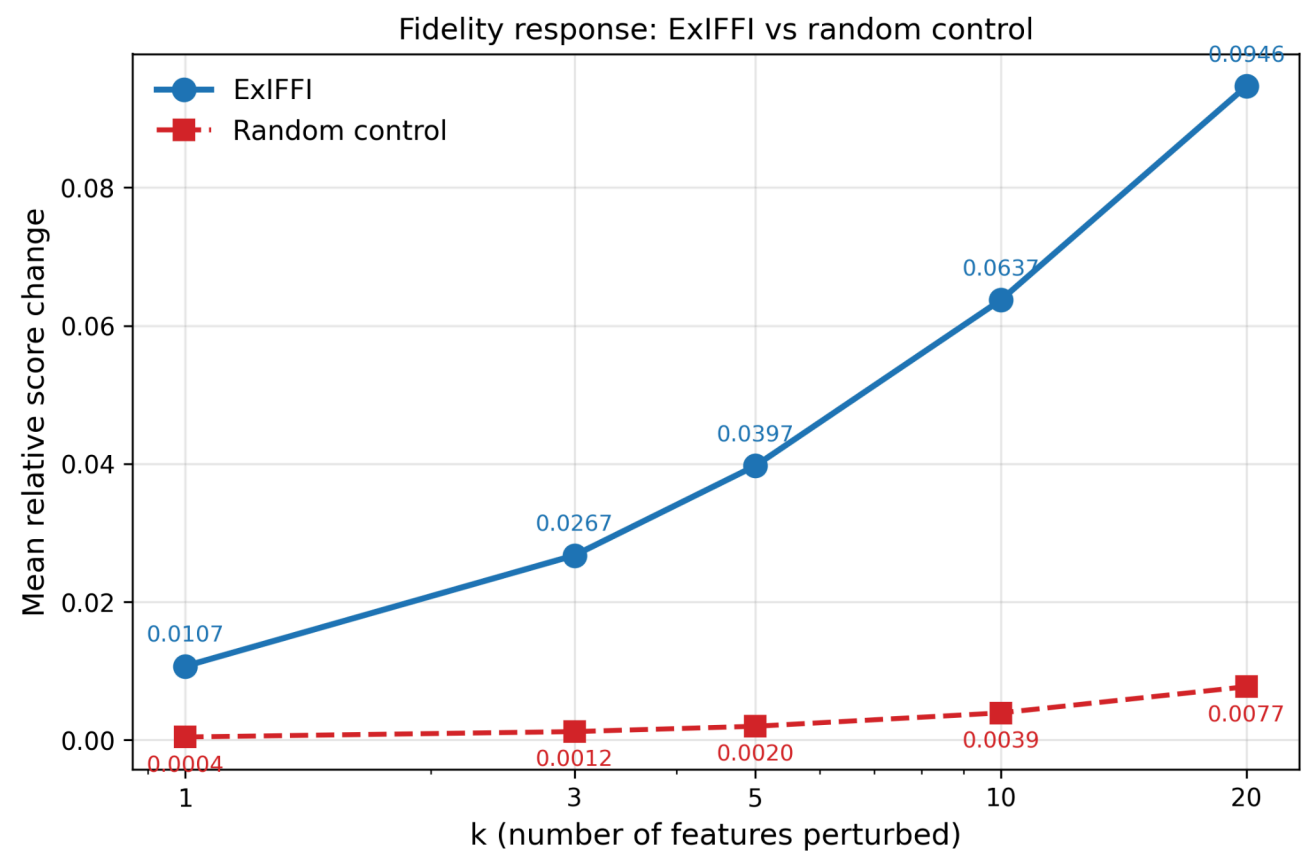

*Figure 5.4.* *Fidelity response curves*

The magnitude of the ExIFFI-to-random ratio is the primary evidence that the metric captures decision relevance, rather than intervention magnitude.
The consistent ExIFFI advantage across all $k$ values and all windows indicate that the features ExIFFI identifies mostly shape the model's output.

## 5.5. Temporal Fidelity Trajectories

While the structural composition of feature attribution across windows is consistent and the fidelity of the explanations has been validated across top-k values, this section investigates whether data drift induces systematic fidelity degradation across evaluation windows.
Two comparisons are considered. The first measures each model's fidelity against its own decisions, tracking how a model's explanations hold up as it faces progressively later windows. The second measures whether the explanation of a previous model is still valid for the current model's output.
The results on within-model fidelity show that all models show lower end-of-horizon fidelity than at W1, although several retrained models exhibit non-monotonic intermediate trajectories, with fidelity at W4 consistently falling below its initial W1 levels across models.
While the magnitude of this overall degradation varies, the downward trajectory across the four temporal windows is universal, demonstrating that no model architecture fully preserves explanation fidelity under data drift.

**Table 5.5.1. Fidelity trajectories per model at k=5**

| Model | W1 | W2 | W3 | W4 | % Change |
|---|---|---|---|---|---|
| **Static** | 0.0349 | 0.0337 | 0.0327 | 0.0317 | −9.1% |
| **Retrained_W1** | 0.0328 | 0.0308 | 0.0306 | 0.0293 | −10.7% |
| **Retrained_W2** | 0.0335 | 0.029 | 0.0308 | 0.0304 | −9.2% |
| **Retrained_W3** | 0.0314 | 0.0286 | 0.0295 | 0.0303 | −3.5% |
| **Retrained_W4** | 0.0306 | 0.029 | 0.0290 | 0.0289 | −5.6% |

Despite this shared drop, individual models exhibit distinct trajectory patterns. The static baseline and W1-retrained models degrade monotonically across successive windows, while models retrained on later windows (W2, W3, and W4) show non-monotonic behaviour, experiencing intermediate recoveries before continuing their downward path.
Though small in magnitude, this fluctuation indicates that explanation fidelity does not decay strictly as a function of temporal distance from the training window.

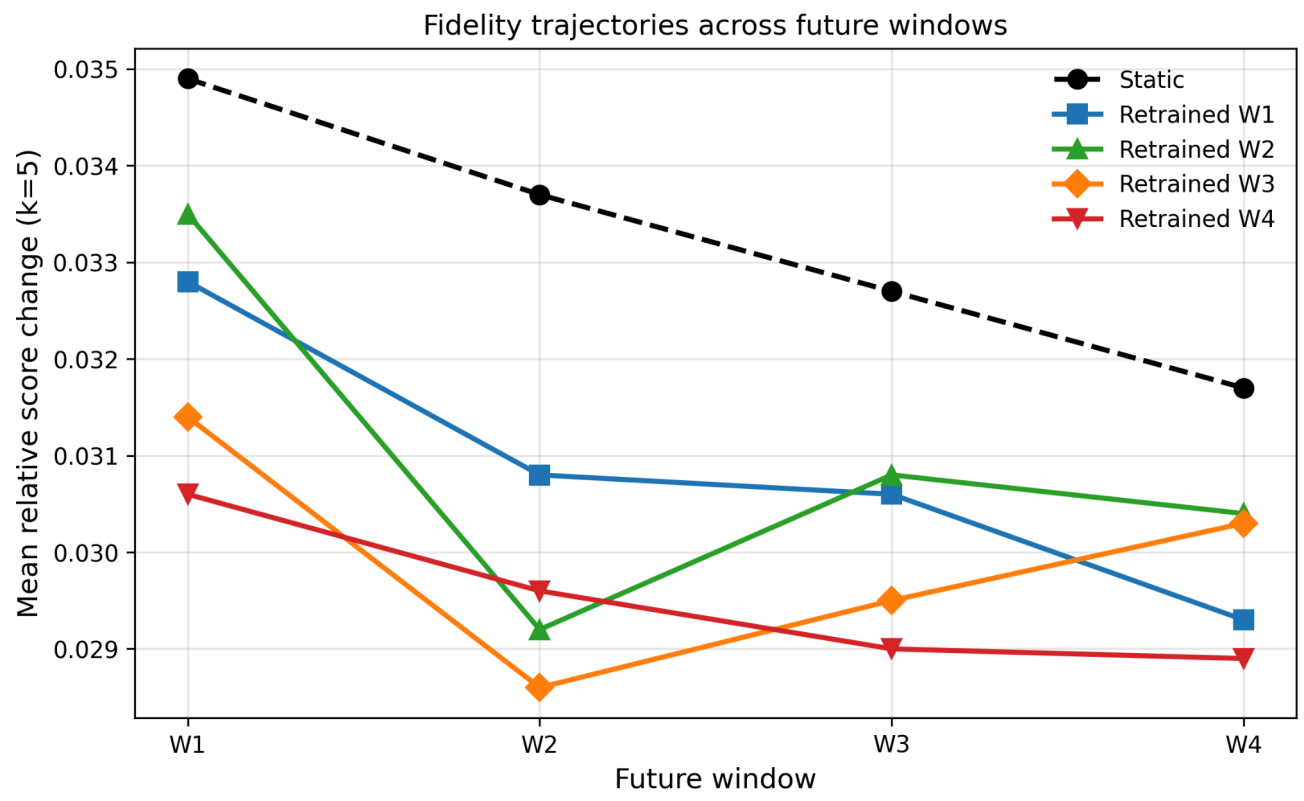

*Figure 5.5.1 Fidelity trajectories across future windows*

Comparing the explanations of the static model against the decisions of the models for each window, retrained on the cumulative data up to its window, shows a consistent and interpretable pattern.
The cumulative model's own explanation produces a larger score change when perturbed than the static explanation, by about 25 to 30 percent.

**Table 5.5.2. Static explanation transfer versus current model baseline at k=5**

| | Base line | Static | Random Control | Static/ Random | Transfer/ Baseline |
|---|---|---|---|---|---|
| W1 | 0.0360 | 0.0270 | 0.0019 | 14.2× | 0.75 |
| W2 | 0.0407 | 0.0291 | 0.0019 | 14.9× | 0.72 |
| W3 | 0.0394 | 0.0303 | 0.0019 | 16.3× | 0.77 |
| W4 | 0.0407 | 0.0291 | 0.0021 | 13.7× | 0.72 |

Three key observations emerge. First, the static explanation's fidelity to the current model is stable across all windows, remaining in a narrow band between 0.0270 and 0.0303.
The explanation does not collapse as drift accumulates.
Secondly, the static model's deficit relative to the current model's own explanation is also stable, with ranges between 0.72 and 0.77. The advantage of the current model does not grow as data drift accumulates, and incurs a fixed cost for using an older explanation, rather than a compounding with each window.
Finally, the static explanation is nowhere near the random control at any window. This establishes that the transferable signal is substantive, not residual.
Taken together, these analyses refine the interpretation of fidelity under drift. Fidelity declines subtly for static models across windows as the model's explanations are evaluated on progressively later data. While explanations produced by an earlier model retain substantial but incomplete transferability to a newer model's decisions.

Therefore, the original explanation does not become invalid; it becomes less precise. Its features remain the most decision relevant features in the current model, but freshly generated explanations capture a larger share of the current decision process.

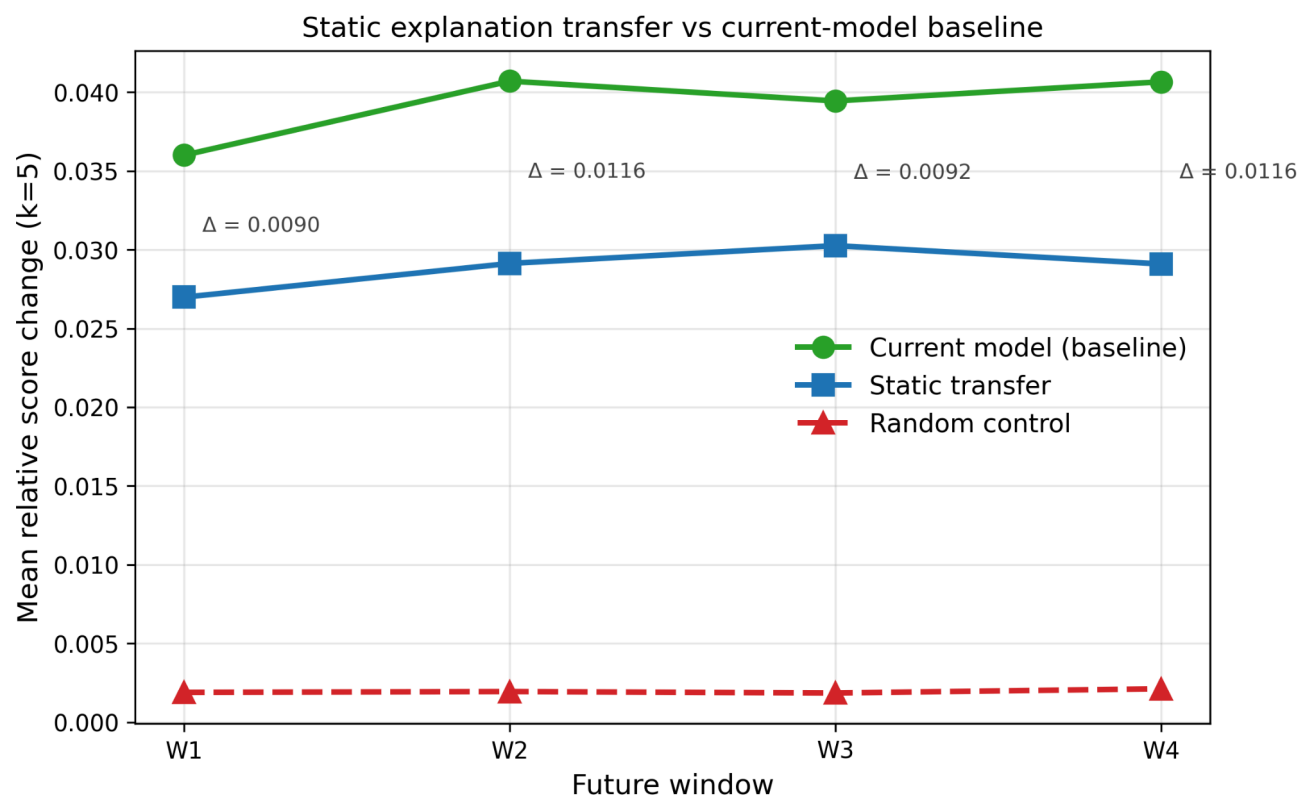

*Figure 5.5.2 Static explanation transfer versus current-model baseline across windows.*

# 6. Discussions

By isolating input distribution shifts from changes in the model, the results support a single, coherent finding: under input distributional shift, the explanation fidelity of a static anomaly detection model degrades, and prior explanations, while still informative, are less faithful to current decisions than freshly generated ones.

This section connects our findings to the broader literature on anomaly detection, explanation faithfulness, and drift adaptation.

## 6.1. Performance Decline Under Data Drift

Every evaluated model showed monotonic decline across the four windows. For the static baseline, F1 falls by 24% to 30% and PR-AUC by 45% to 49% between W1 and W4. The decline is steepest between W2 and W3, coinciding with the largest contraction in fraud prevalence.

First, degradation is not attributable to statistically detectable concept drift, because the drift analysis found no significant change in the conditional relationship $P(Y|X)$, which is the defining distinction between concept drift and changes in the input distribution [9], [18]. The detector degrades because the input distribution has shifted, not because the fraud pattern has changed.

Second, degradation is not attributable to sample size. The static model is trained on 354,324 observations, while each retrained model is trained on 59,054, yet both decline at comparable rates.

If training sample size were the driver, we would expect the static model to degrade more slowly. It does not.

The detection decline is the backdrop against which the fidelity findings must be interpreted.

If fidelity degraded while detection remained stable, the two would be dissociated, and fidelity loss would be a silent failure invisible to detection monitoring.

That is not what we observe. Detection and fidelity degrade together, and the paper's contribution is to characterise how they respond to the same drift under different adaptation strategies.

## 6.2. Stability of Explanation Structure

The findings reveal that while the structural properties of the explanations exhibit systematic, statistically consistent shifts under data drift, the absolute magnitude of these changes remains modest.

Across temporal windows, the primary concentration metric (top-5 mass) decreases by approximately one percentage point, accompanied by a 0.09 nat increase in Shannon entropy and an expansion of 15 to 30 features in effective feature count.

Furthermore, rank correlations between adjacent windows remain high (ranging from 0.81 to 0.83 for the static model), showing that the relative ordering of feature importance is largely preserved under distribution shift.

Even though the gradual rise in entropy and attribution diffuseness suggests that as observations drift into previously unobserved feature-space regions, isolation trees distribute partition depth and attribution weights across a broader set of secondary features, the prevailing rank persistence confirms that explanation structure is relatively insensitive to the magnitude of covariate shift present in this domain.

However, preservation of rank order guarantees neither that features retain their magnitude of predictive influence nor that top-ranked attributions remain faithful to the model's decision process [8], [17].

This decoupling between structural stability and decision relevance directly underlines the necessity of moving beyond surface-level structural metrics toward intervention-based fidelity evaluations.

## 6.3. Temporal Degradation of Fidelity

The validation results show that when the top-k ExIFFI features are replaced with training-period medians, the model's score changes substantially.

The response is monotonic in k, specific to the ExIFFI -selected features versus randomly selected features, and consistent across windows.

At k=1, ExIFFI produces a mean relative score change 23 times larger than the random control; at k=20, the ratio is 12.

The declining ratio as k grows is expected; larger random subsets are more likely to include a decision-relevant feature by chance, but the ExIFFI advantage holds at every k.
This provides empirical evidence that the ExIFFI-selected features have greater decision relevance under the evaluated intervention than randomly selected features.
Such evaluation is consistent with functional approaches to assessing explanation faithfulness [8].
The temporal results, however, show that explanation fidelity declines across windows for every model.
For the static baseline, fidelity falls by 9.1% between W1 and W4. The finding that fidelity degrades means that a fixed decision function is not sufficient to preserve explanation faithfulness and the measured decision-level fidelity under the specified top(k) median-substitution intervention changes across windows..
This constitutes the paper's central empirical finding and motivates the distinction between explanation structure and decision-level fidelity.

### 6.4. Implication for Explanation Transferability

The fourth research question considers whether retraining on drifted distribution preserves explanation fidelity better than leaving the model static, reflecting the broader use of model adaptation as a response to changing data distributions [9].
Results show that while features identified by an explanation generated before drift still matter to the current model's decisions, the static explanation consistently falls below the current model's own explanation by roughly 25 to 30 percent.
This provides evidence that a freshly generated explanation captures a larger share of the current model's decision process than a transferred one does, and this advantage persists across the entire evaluation grid.
The fidelity ratio between the static model and the retrained models remains stable between 0.72 and 0.77 across windows, with no trend toward widening or narrowing over time. Within the four evaluated windows, the transferability gap remained within a relatively narrow range, providing no evidence of a systematic widening of the gap over time.
Therefore, under drift, the static model's explanations remain partially relevant to whatever model is currently making decisions, but their relevance falls short of fully capturing the current decision logic.

### 6.5. Implication for Anomaly Detection Systems

The findings have three direct implications for developing and deploying anomaly detection systems.
First, detection monitoring alone is not sufficient to establish that a model's explanations remain trustworthy. A model can continue to produce reasonable predictions while the features its explanations identify become less representative of the decisions being made.
Secondly, explanation fidelity is not preserved indefinitely. While a static model's explanations remain informative, they are consistently less faithful to the current decisions than freshly generated explanations.
Thirdly, under the drift regime evaluated, the transferability gap between stale and current explanations remained stable, with no trend toward either widening or narrowing.
This means that when operational requirements demand accurate descriptions of current decision logic, regenerating explanations after every model update is the most defensible practice. When explanations are used diagnostically or as a secondary signal alongside detection metrics, transferred explanations retain enough signal to remain informative.
The choice between these uses is a policy decision, but the empirical fact is that transferred and freshly generated explanations are not interchangeable, and the difference does not disappear over time.

## 7. Limitations

A few scope and methodological constraints bound the generalizability of these findings. Primary among these limitations is the scope of the experimental pipeline, which evaluates a single fraud dataset (IEEE-CIS), one detector architecture (Isolation Forest), and one specific attribution framework (ExIFFI).
Consequently, the observed trajectories reflect empirical evidence for this specific configuration, rather than universal assertions, applicable across arbitrary anomaly detection models or drift environments.
Also, our decision-level faithfulness evaluation relies on measuring relative score changes following top-k median substitution. While alternative perturbation strategies, such as mean imputation, out-of-distribution sampling, or feature masking, would likely yield different response magnitudes, the relative advantage of targeted ExIFFI attributions over random baselines is expected to hold across schemes.
Similarly, the temporal evaluation is bounded by its discrete evaluation windows. Although four windows may suffice to establish directional trends and trajectory stability, they represent a finite sequence. Therefore claims regarding the stability of the transferability gap remain constrained by this timeframe.

## 8. Conclusion

The study's central finding is that under data drift, detection models experience decline in explanation fidelity, and that regenerating explanations after a model update recovers part of the lost fidelity.
The broader implication is that fidelity is a property of the deployed model that warrants its own monitoring.
Detection metrics do not detect fidelity loss, and structural stability does not imply functional fidelity. Only a decision -level test exposes the difference. The intervention-based metric used in this study is one concrete way to conduct such a test, and its validation on the IEEE-CIS dataset establishes that it produces a stable and interpretable signal across windows.

The study also established that while original explanations retain substantial usefulness under data drift, as its features are consistently related to the ones the current model relies on and their relationship with the target is still preserved, the explanation fidelity becomes less precise in accurately reflecting current decisions logic.

In practice, this argues for treating explanations as artifacts tied to the model that produced them, and for regenerating them whenever the model is updated, while recognising that stale explanations still carry decision relevant signals and can serve as secondary diagnostics during the interval between updates.

## References

[1] Aggarwal, C.C. (2017) Outlier Analysis. 2nd edn. Cham: Springer.

[2] Arcudi, A., Frizzo, D., Masiero, C. and Susto, G.A. (2023) “Enhancing interpretability and generalizability in Extended Isolation Forests”, *arXiv preprint, arXiv*:2310.05468.

[3] Bhatt, U., Weller, A. and Moura, J.M.F. (2020) “Evaluating and Aggregating Feature-based Model Explanations”, *Proceedings of the Twenty-Ninth International Joint Conference on Artificial Intelligence (IJCAI-20)*, pp. 3010–3016. https://doi.org/10.24963/ijcai.2020/417

[4] Benjamini, Y. and Hochberg, Y. (1995) “Controlling the false discovery rate: A practical and powerful approach to multiple testing”, *Journal of the Royal Statistical Society: Series B, 57(1)*, pp. 289–300.

[5] Carletti, M., Terzi, M. and Susto, G.A. (2023) “Interpretable anomaly detection with DIFFI: Depth-based Isolation Forest feature importance”, *Engineering Applications of Artificial Intelligence*, 119, 105791. https://doi.org/10.1016/j.engappai.2022.105730

[6] Chandola, V., Banerjee, A. and Kumar, V. (2009) “Anomaly detection: A survey”, *ACM Computing Surveys, 41(3)*, Article 15, pp. 1–58. https://doi.org/10.1145/1541880.1541882

[7] Dal Pozzolo, A., Caelen, O., Johnson, R.A. and Bontempi, G. (2018) “Credit card fraud detection: A realistic modeling and a novel learning strategy”, *IEEE Transactions on Neural Networks and Learning Systems,* 29(8), pp. 3784–3797. https://doi.org/10.1109/TNNLS.2017.2736643

[8] Doshi-Velez, F. and Kim, B. (2017) “Towards a rigorous science of interpretable machine learning”, *arXiv preprint, arXiv*:1702.08608. https://doi.org/10.48550/arXiv.1702.08608

[9] Gama, J., Žliobaitė, I., Bifet, A., Pechenizkiy, M. and Bouchachia, A. (2014) “A survey on concept drift adaptation”, *ACM Computing Surveys*, 46(4), Article 44, pp. 1–37. https://doi.org/10.1145/2523813

[10] Hariri, S., Kind, M.C. and Brunner, R.J. (2021) “Extended Isolation Forest”, *IEEE Transactions on Knowledge and Data Engineering*, 33(4), pp. 1479–1489. https://doi.org/10.1109/TKDE.2019.2947676

[11] Hodge, V.J. and Austin, J. (2004) “A survey of outlier detection methodologies”, *Artificial Intelligence Review*, 22, pp. 85–126.

[12] IEEE Computational Intelligence Society (2019) “IEEE-CIS Fraud Detection dataset”, Kaggle. Available at: https://www.kaggle.com/c/ieee-fraud-detection.

[13] Li, Z., Zhu, Y. and Van Leeuwen, M. (2023) “A survey on explainable anomaly detection”, *ACM Transactions on Knowledge Discovery from Data*, 18(1). https://doi.org/10.1145/3609333

[14] Liu, F.T., Ting, K.M. and Zhou, Z.-H. (2008) “Isolation Forest”, *Eighth IEEE International Conference on Data Mining*. IEEE, pp. 413–422. https://doi.org/10.1145/3609333

[15] Lopez-Paz, D. and Oquab, M. (2017) “Revisiting classifier two-sample tests”, in *International Conference on Learning Representations (ICLR 2017*). https://doi.org/10.48550/arXiv.1610.06545

[16] Lundberg, S.M. and Lee, S.-I. (2017) “A unified approach to interpreting model predictions”, in *Advances in Neural Information Processing Systems* 30 (NeurIPS 2017), pp. 4765–4774. https://doi.org/10.48550/arXiv.1705.07874

[17] Miller, T. (2019) “Explanation in artificial intelligence: Insights from the social sciences”, *Artificial Intelligence*, 267, pp. 1–38. https://doi.org/10.1016/j.artint.2018.07.007

[18] Moreno-Torres, J.G., Raeder, T., Alaiz-Rodríguez, R., Chawla, N.V. and Herrera, F. (2012) "A unifying view on dataset shift in classification", *Pattern Recognition*, 45(1), pp. 521–530. https://doi.org/10.1016/j.patcog.2011.06.019

[19] Ribeiro, M.T., Singh, S. and Guestrin, C. (2016) "Why should I trust you?: Explaining the predictions of any classifier", in *Proceedings of the 22nd ACM SIGKDD International Conference on Knowledge Discovery and Data Mining. New York: ACM*, pp. 1135–1144. https://doi.org/10.1145/2939672.2939778

[20] Ruff, L., Kauffmann, J.R., Vandermeulen, R.A., Montavon, G., Samek, W., Kloft, M., Dietterich, T.G. and Müller, K.-R. (2021) "A unifying review of deep and shallow anomaly detection", *Proceedings of the IEEE*, 109(5), pp. 756–795. https://doi.org/10.1109/JPROC.2021.3052449

[21] Saito, T. and Rehmsmeier, M. (2015) "The precision-recall plot is more informative than the ROC plot when evaluating binary classifiers on imbalanced datasets", *PLoS ONE*, 10(3), e0118432. https://doi.org/10.1371/journal.pone.0118432

[22] Shimodaira, H. (2000) 'Improving predictive inference under covariate shift by weighting the log-likelihood function', Journal of Statistical Planning and Inference, 90(2), pp. 227–244. https://doi.org/10.1016/S0378-3758(00)00115-4

[23] Yeh, C.-K., Hsieh, C.-Y., Suggala, A.S., Inouye, D.I. and Ravikumar, P.K. (2019) "On the (In)fidelity and Sensitivity of Explanations", *Advances in Neural Information Processing Systems 32 (NeurIPS 2019)*, pp. 10967–10978.

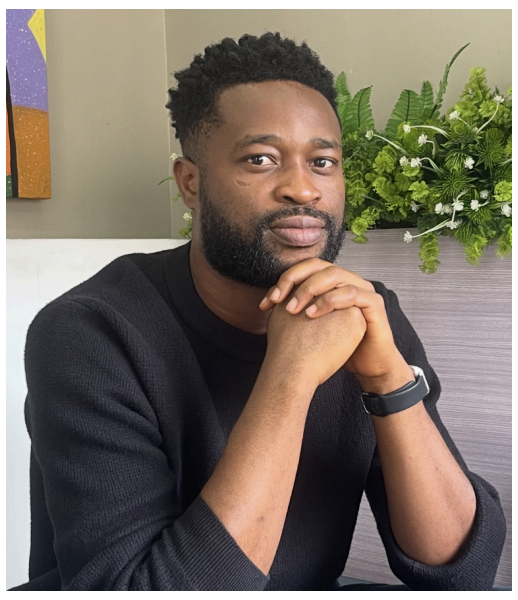

**Samuel Ozechi** is currently a Master's student in Computer Science at the University of East London, UK, and works as an AI/ML Engineer at Interswitch Group. His research interests focus on AI, machine learning, and intelligent systems, with an emphasis on developing adaptive, robust, efficient, and explainable anomaly detection systems for threat intelligence. He is particularly interested in bridging research and practice by exploring how advances in AI and machine learning can inform the development of reliable, value -driven systems for real-world applications.